# Scalable Matrix-valued Kernel Learning for High-dimensional Nonlinear Multivariate Regression and Granger Causality

**Vikas Sindhwani**
IBM Research
Yorktown Heights, NY 10598
vsindhw@us.ibm.com

**Hà Quang Minh**
Istituto Italiano di Tecnologia
Genova, 16163, Italy
minh.haquang@iit.it

**Aurélie C. Lozano**
IBM Research
Yorktown Heights, NY 10598
aclozano@us.ibm.com

## Abstract

We propose a general matrix-valued multiple kernel learning framework for high-dimensional nonlinear multivariate regression problems. This framework allows a broad class of mixed norm regularizers, including those that induce sparsity, to be imposed on a dictionary of vector-valued Reproducing Kernel Hilbert Spaces. We develop a highly scalable and eigendecomposition-free algorithm that orchestrates two inexact solvers for simultaneously learning both the input and output components of separable matrix-valued kernels. As a key application enabled by our framework, we show how high-dimensional causal inference tasks can be naturally cast as sparse function estimation problems, leading to novel nonlinear extensions of a class of Graphical Granger Causality techniques. Our algorithmic developments and extensive empirical studies are complemented by theoretical analyses in terms of Rademacher generalization bounds.

## 1 Introduction

Consider the problem of estimating an unknown nonlinear function, $f : \mathcal{X} \mapsto \mathcal{Y}$, from labeled examples, where $\mathcal{Y}$ is a "structured" output space [6]. In principle, $\mathcal{Y}$ may be endowed with a general Hilbert space structure, though we focus on the multivariate regression setting, where $\mathcal{Y} \subseteq \mathbb{R}^n$. Such problems can be naturally formulated as Tikhonov Regularization [35] in a suitable *vector-valued* Reproducing Kernel Hilbert Space (RKHS) [25, 1]. The theory and formalism of vector-valued RKHS can be traced as far back as the work of Laurent Schwarz in 1964 [32]. Yet, vector-valued extensions of kernel methods have not found widespread application, in stark contrast to the versatile popularity of their scalar cousins. We believe that two key factors are responsible:

- The kernel function is much more complicated - it is *matrix-valued* in this setting. Its choice turns into a daunting model selection problem. Contrast this with the scalar case where Gaussian or Polynomial kernels are a default choice requiring only a few hyperparameters to be tuned.
- The associated optimization problems, in the most general case, have much greater computational complexity than in the scalar case. For example, a vector-valued Regularized Least Squares (RLS) solver would require cubic time in the number of samples *multiplied by the number of output coordinates*.

Scalable kernel learning therefore becomes a basic necessity – an unavoidable pre-requisite – for even considering vector-valued RKHS methods for an application at hand. Our contributions in this paper are as follows:

- We propose a general framework for function estimation over a dictionary of vector-valued RKHSs where a broad family of variationally defined regularizers, including sparsity inducing norms, serve to optimally combine a collection of matrix-valued kernels. As such our framework may be viewed as providing generalizations of scalar multiple kernel learning [20, 26, 21, 29] and associated structured sparsity algorithms [8].
- We specialize our framework to the class of *separable kernels* [1] which are of interest due to their universality [9], conceptual simplicity and potential for scalability. Separable matrix-valued kernels are composed of a *scalar input kernel* component and a *positive semi-definite output matrix* component (to be formally defined later). We provide a full resolution of the kernel learning problem in this setting by jointly estimating both components together. This is in contrast to recent efforts [12, 18] where only one of the two components is optimized, and the

- full complexity of the joint problem is not addressed. Our algorithms achieve scalability by orchestrating carefully designed inexact solvers for inner subproblems, for which we also provide convergence rates.
- We provide bounds on Rademacher Complexity for the vector-valued hypothesis sets considered by our algorithm. This complements and extends generalization results in the multiple kernel learning literature for the scalar case [11, 38, 19, 24, 20].
- We demonstrate that the generality of our framework enables novel non-standard applications. In particular, when applied to multivariate time series problems, sparsity in kernel combinations lends itself to a natural causality interpretation. We believe that this nonlinear generalization of graphical Granger Causality techniques (see [3, 22, 33] and references therein) may be of independent interest.

A number of results are stated in this paper for completeness, but without proofs due to space considerations. For detailed proofs, please see a version of this paper with Supplementary Material available at [34].

## 2 Vector-valued RLS & Separable Matrix-valued Kernels

Given labeled examples $\{(\mathbf{x}_i, \mathbf{y}_i)\}_{i=1}^l$, $\mathbf{x}_i \in \mathcal{X} \subset \mathbb{R}^d$, $\mathbf{y}_i \in \mathcal{Y} \subset \mathbb{R}^n$, the vector-valued Regularized Least Squares (RLS) solves the following problem,

$$\underset{f \in \mathcal{H}_{\overrightarrow{k}}}{\arg\min} \frac{1}{l} \sum_{i=1}^l \|f(\mathbf{x}_i) - \mathbf{y}_i\|_2^2 + \lambda \|f\|_{\mathcal{H}_{\overrightarrow{k}}}^2, \quad (1)$$

where $\mathcal{H}_{\overrightarrow{k}}$ is a vector-valued RKHS generated by the kernel function $\overrightarrow{k}$, and $\lambda > 0$ is the regularization parameter. For readers unfamiliar with vector-valued RKHS theory that is the basis of such algorithms, we provide a first-principles overview in our Supplementary Material (see [34], section 4).

In the vector-valued setting, $\overrightarrow{k}$ is a matrix-valued function, i.e., when evaluated for any pair of inputs $(\mathbf{x}, \mathbf{z})$, the value of $\overrightarrow{k}(\mathbf{x}, \mathbf{z})$ is an $n$-by-$n$ matrix. More generally speaking, the kernel function is an input-dependent linear operator on the output space. The kernel function is *positive* in the sense that for any finite set of $l$ input-output pairs $\{(\mathbf{x}_i, \mathbf{y}_i)\}_{i=1}^l$, the following holds: $\sum_{i,j=1}^l \mathbf{y}_i^T \overrightarrow{k}(\mathbf{x}_i, \mathbf{x}_j) \mathbf{y}_j \geq 0$. A generalization [25] of the standard Representer Theorem says that the optimal solution has the form,

$$f(\cdot) = \sum_{i=1}^l \overrightarrow{k}(\mathbf{x}_i, \cdot) \boldsymbol{\alpha}_i, \quad (2)$$

where the coefficients $\boldsymbol{\alpha}_i$ are $n$-dimensional vectors.

For RLS, these coefficient vectors can be obtained solving a dense linear system, of the familiar form,

$$\left(\overrightarrow{\mathbf{K}} + \lambda l \mathbf{I}_{nl}\right) vec(\mathbf{C}^T) = vec(\mathbf{Y}^T),$$

where $\mathbf{C} = [\boldsymbol{\alpha}_1 \ldots \boldsymbol{\alpha}_l]^T \in \mathbb{R}^{l \times n}$ assembles the coefficient vectors into a matrix; the *vec* operator stacks columns of its argument matrix into a long column vector; $\overrightarrow{\mathbf{K}}$ is a large $nl \times nl$-sized Gram matrix comprising of the blocks $\overrightarrow{k}(\mathbf{x}_i, \mathbf{x}_j)$, for $i, j = 1 \ldots l$, and $\mathbf{I}_{nl}$ denotes the identity matrix of compatible size. It is easy to see that for $n = 1$, the above developments exactly collapse to familiar concepts for scalar RLS (also known as Kernel Ridge Regression). In general though, the above linear system requires $O((nl)^3)$ time to be solved using standard dense numerical linear algebra, which is clearly prohibitive. However, for a family of *separable matrix-valued kernels* [1, 9] defined below, the computational cost can be improved to $O(n^3 + l^3)$; though still costly, this is atleast comparable to scalar RLS when $n$ is comparable to $l$.

**Separable Matrix Valued Kernel and its Gram matrix**: *Let $k$ be a scalar kernel function on the input space $\mathcal{X}$ and $\mathbf{K}$ represent its Gram matrix on a finite sample. Let $\mathbf{L}$ be an $n \times n$ positive semi-definite output kernel matrix. Then, the function $\overrightarrow{k}(\mathbf{x}, \mathbf{z}) = k(\mathbf{x}, \mathbf{z}) \mathbf{L}$ is positive and hence defines a matrix valued kernel. The Gram matrix of this kernel is $\overrightarrow{\mathbf{K}} = \mathbf{K} \otimes \mathbf{L}$ where $\otimes$ denotes Kronecker product.*

For separable kernels, the corresponding RLS dense linear system (Eqn 3 below) can be reorganized into a *Sylvester equation* (Eqn 4 below):

$$(\mathbf{K} \otimes \mathbf{L} + \lambda l \mathbf{I}_{nl}) vec(\mathbf{C}^T) = vec(\mathbf{Y}^T), \quad (3)$$
$$\mathbf{KCL} + \lambda l \mathbf{C} = \mathbf{Y}. \quad (4)$$

Sylvester solvers are more efficient than applying a direct dense linear solver for Eqn 3. The classical Bartel-Stewart and Hessenberg-Schur methods (e.g., see MATLAB's *dlyap* function) are usually used for solving Sylvester equations. They are similar in flavor to an eigendecomposition approach [1] we describe next for completeness, though they take fewer floating point operations at the same cubic order of complexity.

**Eigen-decomposition based Sylvester Solver:** Let $\mathbf{K} = \mathbf{TMT}^T$ and $\mathbf{L} = \mathbf{SNS}^T$ denote the eigendecompositions of $\mathbf{K}$ and $\mathbf{L}$ respectively, where $\mathbf{M} = diag(\sigma_1 \ldots \sigma_l), \mathbf{N} = diag(\rho_1 \ldots \rho_n)$. Then the solution to the matrix equation $\mathbf{KCL} + \lambda \mathbf{C} = \mathbf{Y}$ always exists when $\lambda > 0$ and is given by $\mathbf{C} = \mathbf{T\tilde{X}S}$ where $\tilde{\mathbf{X}}_{ij} = \frac{(\mathbf{T}^T \mathbf{YS})_{ij}}{\sigma_i \rho_j + \lambda}$.

**Output Kernel Learning**: In recent work [12] develop an elegant extension of the vector-valued RLS

problem (Eqn. 1), which we will briefly describe here. We will use the shorthand $\vec{k} = k\mathbf{L}$ to represent the implied separable kernel and correspondingly denote its RKHS by $\mathcal{H}_{k\mathbf{L}}$. [12] attempt to jointly learn both $f \in \mathcal{H}_{k\mathbf{L}}$ and $\mathbf{L}$, *for a fixed pre-defined choice of k*. In finite dimensional language, $\mathbf{L}$ and $\mathbf{C}$ are estimated by solving the following problem [12],

$$\underset{\mathbf{C} \in \mathbb{R}^{l \times n}, \mathbf{L} \in \mathcal{S}_+^n}{\arg\min} \frac{1}{l}\|\mathbf{KCL}-\mathbf{Y}\|_F^2 + \lambda tr(\mathbf{C}^T\mathbf{KCL}) + \rho\|\mathbf{L}\|_F^2,$$

where $tr(\cdot)$ denotes trace, $\|\cdot\|_F$ denotes Frobenius norm, and $\mathcal{S}_+^n$ denotes the cone of positive semi-definite matrices. It is shown that the objective function is *invex*, i.e., its stationary points are globally optimal. [12] proposed a block coordinate descent where for fixed $\mathbf{L}$, $\mathbf{C}$ is obtained by solving Eqn. 4 using an Eigendecomposition-based solver. Under the assumption that $\mathbf{C}$ *exactly* satisfies Eqn. 4, the resulting update for $\mathbf{L}$ is then shown to automatically satisfy the constraint that $\mathbf{L} \in \mathcal{S}_+^n$. However, [12] remark that experiments on their largest dataset took roughly a day to complete on a standard desktop and that the *"limiting factor was the solution of the Sylvester equation"*.

## 3 Learning over a Vector-valued RKHS Dictionary

Our goals are two fold: one, we seek a fuller resolution of the separable kernel learning problem for vector-valued RLS problems; and two, we wish to derive extensible algorithms that are eigendecomposition-free and much more scalable. In this section, we expand Eqn. 1 to simultaneously learn both input and output kernels over a predefined dictionary, and develop optimization algorithms based on approximate inexact solvers that execute cheap iterations.

Consider a **dictionary of separable matrix valued kernels**, of size $m$, sharing the same output kernel matrix $\mathbf{L}$: $\mathcal{D}_\mathbf{L} = \{k_1\mathbf{L}, \ldots k_m\mathbf{L}\}$. Let $\mathcal{H}(\mathcal{D}_\mathbf{L})$ denote the sum space of functions:

$$\mathcal{H}(\mathcal{D}_\mathbf{L}) = \left\{ f = \sum_{j=1}^m f_j : f_j \in \mathcal{H}_{k_j\mathbf{L}} \right\}, \quad (5)$$

and equip this space with the following $l_p$ norms:

$$\|f\|_{l_p(\mathcal{H}(\mathcal{D}_\mathbf{L}))} = \inf_{f: f=\sum_j f_j} \left\| (\|f_1\|_{\mathcal{H}_{k_1\mathbf{L}}}, \ldots, \|f_m\|_{\mathcal{H}_{k_m\mathbf{L}}}) \right\|_p.$$

The infimum in the above definition is in fact attained as a minimum (see Proposition 2 in our Supplementary Material [34]), so that one can write

$$\|f\|_{l_p(\mathcal{H}(\mathcal{D}_\mathbf{L}))} = \min_{f: f=\sum_j f_j} \left\| (\|f_1\|_{\mathcal{H}_{k_1\mathbf{L}}}, \ldots, \|f_m\|_{\mathcal{H}_{k_m\mathbf{L}}}) \right\|_p.$$

For notational simplicity, we will denote these norms as $\|f\|_{l_p}$ though it should be kept in mind that their definition is with respect to a given dictionary of vector-valued RKHSs.

Note that $\|f\|_{l_1}$, being the $l_1$ norm of the vector of norms in individual RKHSs, imposes a *functional notion of sparsity* on the vector-valued function $f$. We now consider objective functions of the form,

$$\underset{f \in \mathcal{H}(\mathcal{D}_\mathbf{L}), \mathbf{L} \in \mathcal{S}_+^n(\tau)}{\arg\min} \frac{1}{l}\sum_{i=1}^l \|f(\mathbf{x}_j) - \mathbf{y}_i\|_2^2 + \lambda\Omega(f), \quad (6)$$

where $\mathbf{L}$ is constrained to belong to the *Spectahedron* with bounded trace:

$$\mathcal{S}_+^n(\tau) = \{\mathbf{X} \in \mathcal{S}_+^n | trace(\mathbf{X}) \leq \tau\},$$

where $\mathcal{S}_+^n$ denotes the cone of symmetric positive semi-definite matrices, and $\Omega$ is a regularizer whose canonical choice will be the squared $l_p$ norm (with $1 \leq p \leq 2$), i.e.,

$$\Omega(f) = \|f\|_{l_p(\mathcal{H}(\mathcal{D}_\mathbf{L}))}^2.$$

When $p \to 1$, $\Omega$ induces sparsity, while for $p \to 2$, non-sparse uniform combinations approaching a simple sum of kernels is induced. Our algorithms work for a broad choice of regularizers that admit a quadratic variational representation of the form:

$$\Omega(f) = \min_{\boldsymbol{\eta} \in \mathbb{R}_+^m} \sum_{i=1}^m \frac{\|f_i\|_{\mathcal{H}_{k_i\mathbf{L}}}^2}{\eta_i} + \omega(\boldsymbol{\eta}), \quad (7)$$

for an appropriate auxiliary function $\omega : \mathbb{R}_+^m \mapsto \mathbb{R}$. For squared $l_p$ norms, this auxillary function is the indicator function of a convex set [5, 36, 26]:

$$\omega(\boldsymbol{\eta}) = 0 \text{ if } \eta_i \geq 0, \sum_{i=1}^m \eta_i^q \leq 1, \text{ and } \infty \text{ otherwise}, \quad (8)$$

where $q = \frac{p}{2-p} \in [1, \infty]$ for $p \in [1, 2]$.

We rationalize this framework as follows:

○ Penalty functions of the form above define a broad family of structured sparsity-inducing norms that have extensively been used in the multiple kernel learning and sparse modeling literature [5, 36, 27]. They allow complex non-differentiable norms to be related back to weighted $l_2$ or RKHS norms, and optimizing the weights $\boldsymbol{\eta}$ in many cases infact admits closed form expressions. Infact, all norms admit quadratic variational representations of related forms [5].

○ Optimizing $\mathbf{L}$ over the Spectahedron allows us to develop a specialized version of the approximate Sparse SDP solver [16] whose iterations involve the computation of only a single extremal eigenvector of the (partial) gradient at the current iterate – this involves relatively cheap operations followed by quick rank-one updates.
○ By bounding the trace of **L**, we show below that a Conjugate Gradient (CG) based iterative Sylvester solver for Eqn. 3 would always be invoked on well-conditioned instances and hence show rapid numerical convergence (particularly also with warm starts).
○ The trace constraint parameter $\tau$, together with the regularization parameter $\lambda$, also naturally appears in our Rademacher complexity bounds.

### 3.1 Algorithms

First we give a basic result concerning sums of vector-valued RKHSs. The proof, given in our Supplementary Material [34], follows Section 6 of [4] replacing scalar concepts with corresponding notions from the theory of vector-valued RKHSs [25].

**Proposition 1.** *Given a collection of matrix-valued reproducing kernels $\vec{k}_1 \ldots \vec{k}_m$ and positive scalars $\eta_j > 0, j = 1 \ldots m$, the function:*

$$\vec{k}_{\boldsymbol{\eta}} = \sum_{i=1}^{m} \eta_i \vec{k}_i,$$

*is the reproducing kernel of the sum space $\mathcal{H} = \{f : \mathcal{X} \mapsto \mathcal{Y} | f(\mathbf{x}) = \sum_{j=1}^{m} f_j(\mathbf{x}), f_j \in \mathcal{H}_{\vec{k}_j}\}$ with the norm given by:*

$$\|f\|^2_{\mathcal{H}_{\vec{k}_{\boldsymbol{\eta}}}} = \min_{f = \sum_{j=1}^{m} f_i, f_j \in \mathcal{H}_{\vec{k}_j}} \sum_{j=1}^{m} \frac{\|f_j\|^2}{\eta_j}.$$

This result combined with the variational representation of the penalty function in Eqn. 7 allows us to reformulate Eqn. 6 in terms of a joint optimization problem over $\boldsymbol{\eta}, \mathbf{L}$ and $f \in \mathcal{H}_{k_{\boldsymbol{\eta}}\mathbf{L}}$, where we define the weighted scalar kernel $k_{\boldsymbol{\eta}} = \sum_{j=1}^{m} \eta_j k_j$. This formulation allows us to scale gracefully with respect to $m$, the number of kernels. Denote the Gram matrix of $k_{\boldsymbol{\eta}}$ on the labeled data as $\mathbf{K}_{\boldsymbol{\eta}}$, i.e., $\mathbf{K}_{\boldsymbol{\eta}} = \sum_{j=1}^{m} \eta_j \mathbf{K}_j$, where $\mathbf{K}_j$ denotes the Gram matrices of the individual scalar kernel $k_j$. The finite dimensional version of the reformulated problem becomes,

$$\arg\min_{\mathbf{C} \in \mathbb{R}^{n \times l}, \mathbf{L} \in \mathcal{S}^n_+(\tau), \boldsymbol{\eta} \in \mathbb{R}^m_+} \frac{1}{l} \|\mathbf{K}_{\boldsymbol{\eta}}\mathbf{C}\mathbf{L} - \mathbf{Y}\|^2_F$$
$$+ \lambda \, trace\left(\mathbf{C}^T \mathbf{K}_{\boldsymbol{\eta}} \mathbf{C} \mathbf{L}\right) + \omega(\boldsymbol{\eta}). \qquad (9)$$

A natural strategy for such a non-convex problem is Block Coordinate Descent. The minimization of **C** or **L** keeping the other variables fixed, is a convex optimization problem. The minimization of $\boldsymbol{\eta}$ admits closed form solution. At termination, the vector-valued function returned is

$$f^\star(\mathbf{x}) = \mathbf{L}\mathbf{C}^T[k_{\boldsymbol{\eta}}(\mathbf{x}, \mathbf{x}_1) \ldots k_{\boldsymbol{\eta}}(\mathbf{x}, \mathbf{x}_l)]^T,$$

which is a matrix version of the functional form for the optimal solution as specified by the Representer theorem (Eqn. 2) for separable kernels. We next describe each of the three block minimization subproblems.

**A. Conjugate Gradient Sylvester Solver**: For fixed $\boldsymbol{\eta}, \mathbf{L}$, the optimal **C** is given by the solution of the dense linear system of Eqn 3 or the Sylvester equation 4, with $\mathbf{K} = \mathbf{K}_{\boldsymbol{\eta}}$. General dense linear solvers have prohibitive $O(n^3 l^3)$ cost when invoked on Eqn. 3. The $O(n^3 + l^3)$ eigendecomposition-based Sylvester solver performs much better, but needs to be invoked repeatedly since **L** as well as $\mathbf{K}_{\boldsymbol{\eta}}$ are changing across (outer) iterations. Instead, we apply a CG-based iterative solver for Eqn 3. Despite the massive size of the $nl \times nl$ linear system, using CG infact has several unobvious quantifiable advantages due to the special Kronecker structure of Eqn. 3:

○ A CG solver can exploit warm starts by initializing from previous $\boldsymbol{\eta}, \mathbf{L}$, and allow early termination at cheaper computational cost.
○ The large $nl \times nl$ coefficient matrix in Eqn.3 never needs to be explicitly materialized. For any CG iterate $\mathbf{C}^{(k)}$, matrix-vector products can be efficiently computed since,

$$(\mathbf{K}_{\boldsymbol{\eta}} \otimes \mathbf{L} + \lambda l \mathbf{I}_{nl})vec(\mathbf{C}^{(k)T}) = vec(\mathbf{K}_{\boldsymbol{\eta}} \mathbf{C}^{(k)} \mathbf{L} + \lambda l \mathbf{C}^{(k)}).$$

CG can exploit additional low-rank or sparsity structure in $\mathbf{K}_{\boldsymbol{\eta}}$ and **L** for fast matrix multiplication. When the base kernels are either (a) linear kernels derived from a small group of features, or (b) arise from randomized approximations, such as the random Fourier features for Gaussian Kernel [28], then $\mathbf{K}_{\boldsymbol{\eta}} = \sum_{j=1}^{m} \eta_j \mathbf{Z}_j \mathbf{Z}_j^T$ where $\mathbf{Z}_j$ has $d_j \ll l$ columns. In this case, $\mathbf{K}_{\boldsymbol{\eta}}$ need never be explicitly materialized and the cost of matrix multiplication can be further reduced.

CG is expected to make rapid progress in a few iterations in the presence of strong regularization as enforced jointly by $\lambda$ and the trace constraint parameter $\tau$ on **L**. This is because the coefficient matrix of the linear system in Eqn. 3 is then expected to be well conditioned for all possible $\mathbf{K}_{\boldsymbol{\eta}}, \mathbf{L}$ that the algorithm may encounter, as we formalize in the following proposition. Below, let $\|\mathbf{K}_i\|_2$ denote the spectral norm of the Gram matrix $\mathbf{K}_i$, i.e., its largest eigenvalue.

**Proposition 2** (Convergence Rate for CG-solver for Eqn. 3 with $\mathbf{K} = \mathbf{K}_{\boldsymbol{\eta}}$). *Assume $l_1$ norm for $\Omega$ in Eqn. 6. Let $\mathbf{C}^{(k)}$ be the CG iterate at step $k$, $\mathbf{C}^\star$ be* the optimal solution (at current fixed $\boldsymbol{\eta}$ and $\mathbf{L}$) and $\mathbf{C}^{(0)}$ be the initial iterate (warm-started from previous value). Then,

$$\|\mathbf{C}^{(k)} - \mathbf{C}^*\|_F \leq 2\sqrt{\phi}\left(\frac{\sqrt{\phi}-1}{\sqrt{\phi}+1}\right)^k \|\mathbf{C}^{(0)} - \mathbf{C}^*\|_F,$$

where $\phi = 1 + \frac{\gamma\tau}{l\lambda}$ with $\gamma = \max_i \|\mathbf{K}_i\|_2$. For dictionaries involving only Gaussian scalar kernels, $\phi \leq 1 + \frac{\tau}{\lambda}$, i.e., the convergence rate depends only on the relative strengths of regularization parameters $\lambda, \tau$.

The proof is given in our Supplementary Material [34].

**B. Updates for $\boldsymbol{\eta}$**: Note from Eqn. 7 that the optimal weight vector $\boldsymbol{\eta}$ only depends on the RKHS norms of component functions, and is oblivious to the vector-valued, as opposed to scalar-valued, nature of the functions themselves. This is essentially the reason why existing results [5, 36, 26] routinely used in the (scalar) MKL literature can be immediately applied to our setting to get closed form update rules. Define $\alpha_j = \|f_j\|_{k_j \mathbf{L}} = \hat{\eta}_j \sqrt{trace(\mathbf{CK}_j\mathbf{CL})}$ where $\hat{\eta}_j$ refers to previous value of $\eta_j$. The components of the optimal weight vector $\boldsymbol{\eta}$ are given below for two choices of $\Omega$.

○ For $\Omega(f) = \|f\|_{l_p}^2$, the optimal $\boldsymbol{\eta}$ is given by:

$$\eta_j = \alpha_j^{\frac{2}{q+1}} / \left(\sum_{j=1}^m \alpha_j^{\frac{2}{q+1}}\right)^q \quad \text{for } q = \frac{p}{2-p} \quad (10)$$

○ For an elastic net type penalty, $\Omega(f) = (1-\mu)\|f\|_{l_1}^1 + \mu\|f\|_{l_2}^2$, we have $\eta_j = \alpha_j/(1 - \mu + \mu\alpha_j)$.

Several other choices are also infact possible, e.g., see Table 1 in [36], discussion around subquadratic norms in [5] and regularizers for structured sparsity introduced in [27].

**C. Spectahedron Solver**: Here, we consider the $\mathbf{L}$ optimization subproblem, which is:

$$\underset{\mathbf{L} \in \mathcal{S}_+^n(\tau)}{\arg\min}\, g(\mathbf{L}) = \frac{1}{l}\|\mathbf{AL} - \mathbf{Y}\|_{fro}^2 + \lambda trace(\mathbf{B}^T\mathbf{L}), \quad (11)$$

where $\mathbf{A} = \mathbf{K}_{\boldsymbol{\eta}}\mathbf{C}$ and $\mathbf{B} = \mathbf{C}^T\mathbf{A}$. Hazan's Sparse SDP solver [16, 13] based on Frank-Wolfe algorithm [10], can be used for problems of the general form,

$$\mathbf{L}^\star = \underset{\mathbf{L} \in \mathcal{S}_+^n, trace(\mathbf{L})=1}{\arg\min}\, g(\mathbf{L}),$$

where $g$ is a convex, symmetric and differentiable function. It has been successfully applied in matrix completion and collaborative filtering settings [17].

In each iteration, Hazan's algorithm optimizes a linearization of the objective function around the current iterate $\mathbf{L}^{(k)}$, resulting in updates of the form,

$$\mathbf{L}^{(k+1)} = \mathbf{L}^{(k)} + \alpha_k(\boldsymbol{v}_k\boldsymbol{v}_k^T - \mathbf{L}^{(k)}), \quad (12)$$

where $\boldsymbol{v}_k = ApproxEV\left(\nabla g(\mathbf{L}^{(k)}), \frac{C_g}{k^2}\right)$, $\alpha_k = \min\left(1, \frac{2}{k}\right)$ and $C_g$ is a constant which measures the curvature of the graph of $g$ over the Spectahedron. Here, $ApproxEv$ is an approximate eigensolver which when invoked on the gradient of $g$ at the current iterate $\mathbf{L}^{(k)}$ (a positive semi-definite matrix) computes the single eigenvector corresponding to the smallest eigenvalue, only to a prespecified precision. Hazan's algorithm is appealing for us since *each iteration itself tolerates approximations* and the updates pump in rank-one terms. We specialize Hazan's algorithm to our framework as follows (below, note that $\mathbf{A} = \mathbf{K}_{\boldsymbol{\eta}}\mathbf{C}$ and $\mathbf{B} = \mathbf{C}^T\mathbf{A}$):

○ Using **bounded trace constraints**, $trace(\mathbf{L}) \leq \tau$, instead of unit trace is more meaningful for our setting. The following modified updates optimize over $\mathcal{S}_+^n(\tau)$: $\mathbf{L}^{(k+1)} = \mathbf{L}^{(k)} + \alpha_k(\tau\boldsymbol{v}_k\boldsymbol{v}_k^T - \mathbf{L}^{(k)})$, where $\boldsymbol{v}_k$ is reset to the zero vector if the smallest eigenvalue is positive.

○ The **gradient for our objective** is: $\nabla g(\mathbf{L}) = \boldsymbol{G} + \boldsymbol{G}^T - diag(\boldsymbol{G})$ where $\boldsymbol{G} = \lambda\mathbf{B} + 2\mathbf{A}^T\mathbf{AL} - 2\mathbf{A}^T\mathbf{Y}$ and $diag(\cdot)$ assembles the diagonal entries of its argument into a diagonal matrix.

○ Instead of using Hazan's line search parameter $\alpha_k$, we do **exact line search** along the direction $\boldsymbol{P} = \tau\boldsymbol{v}_k\boldsymbol{v}_k^T - \mathbf{L}^{(k)}$ which leads to a closed form expression:

$$\alpha_k = -\frac{trace((\frac{1}{l}\mathbf{AL}^{(k)} - \mathbf{Y})^T\mathbf{AP} + \frac{1}{2}\lambda\mathbf{BP})}{trace(\frac{1}{l}\boldsymbol{P}^T\mathbf{A}^T\mathbf{AP})}.$$

Adapting the analysis of Hazan's algorithm in [13] to our setting, we get the following convergence rate (proof given in our Supplementary Material [34]):

**Proposition 3** (Convergence Rate for optimizing $\mathbf{L}$). *Assume $l_1$ norm for $\Omega$ in Eqn. 6. For $k \geq 16(\tau\gamma)^2/\epsilon$, the iterate in Eqn. 12 satisfies $g(\mathbf{L}^{(k+1)}) - g(\mathbf{L}^\star) \leq \epsilon/2$ where $\gamma = \max_i \|\mathbf{K}_i\|_2$.*

**Remarks**: Note that Propositions 2 and 3 offer convergence rates for the convex optimization subproblems associated with optimizing $\mathbf{C}$ and $\mathbf{L}$ respectively keeping other variables fixed. These results strongly suggest that inexact solutions to subproblems may be quickly obtained in a few iterations. A full theoretical convergence analysis of Block Coordinate Descent to stationary points of the objective function under inexact updates is currently not within the scope of this paper. An empirical analysis of convergence behavior is provided in Section 4.

## 3.2 Rademacher Complexity Results

Here, we complement our algorithms with statistical generalization bounds. The notion of Rademacher complexity is readily generalizable to vector-valued hypothesis spaces [23]. Let $\mathcal{H}$ be a class of functions $f : \mathcal{X} \to \mathcal{Y}$, where $Y \subset \mathbb{R}^n$. Let $\boldsymbol{\sigma} \in \mathbb{R}^n$ be a vector of independent Rademacher variables, and similarly define the matrix $\Sigma = [\boldsymbol{\sigma}_1, \ldots, \boldsymbol{\sigma}_l] \in \mathbb{R}^{n \times l}$. The empirical Rademacher complexity of the vector-valued class $\mathcal{H}$ is the function $\hat{R}_l(\mathcal{H})$ defined as

$$\hat{\mathcal{R}}_l(\mathcal{H}) = \frac{1}{l}\mathbb{E}_\Sigma \left[\sup_{f \in \mathcal{H}} \sum_{i=1}^{l} \boldsymbol{\sigma}_i^T f(\mathbf{x}_i)\right]. \quad (13)$$

We now state bounds on the Rademacher complexity of hypothesis spaces considered by our algorithms, both for general matrix-valued kernel dictionaries as well as the special case of separable matrix-valued kernel dictionaries. When the output dimensionality is set to 1, our results essentially recover existing results in the *scalar* multiple kernel learning literature given in [11, 38, 19, 24]. Our bounds in part (B) and (C) in the Theorem below involve the same dependence on the number of kernels $m$, and on $p$ (for $l_p$ norms) as given in [11], though there are slight differences in stated bounds since our hypothesis class is not exactly the same as that in [11]. In particular, for the case of $p = 1$ (part C below), we obtain a $\sqrt{\log m}$ dependence on the number of kernels which is known to be tight for the scalar case [11, 24]. Since this logarithmic dependence is rather mild, we can expect to learn effectively over a large dictionary even in the vector-valued setting.

**Theorem 3.1.** *Let $\mathcal{H} = \{f = \sum_{j=1}^{m} f_j, f_j \in \mathcal{H}_{\vec{k}_j}\}$. For $1 \leq p \leq \infty$, consider the hypothesis class*

$$\mathcal{H}_\lambda^p = \{f \in \mathcal{H} : f = \sum_{j=1}^{m} f_j, \quad f_j \in \mathcal{H}_{\vec{k}_j},$$

$$||f||_{\mathcal{H}(l_p)} = \min_{f_j \in \mathcal{H}_{\vec{k}_j}, \sum_{j=1}^{m} f_j = f} \left(\sum_{j=1}^{m} ||f_j||_{\mathcal{H}_{\vec{k}_j}}^p\right)^{1/p} \leq \lambda\}.$$

*(A) For any $p$, $1 \leq p \leq \infty$, the empirical Rademacher complexity of $\mathcal{H}_\lambda^p$ can be upper bounded as follows:*

$$\hat{R}_l(\mathcal{H}_\lambda^p) \leq \frac{\lambda ||\mathbf{u}||_1}{l},$$

*where $\mathbf{u} = \left[\sqrt{\text{trace}(\vec{\mathbf{K}}_1)}, \ldots, \sqrt{\text{trace}(\vec{\mathbf{K}}_m)}\right]$. For the case of separable kernels, where $\vec{k}_i(x, z) = k_i(x, z)\mathbf{L}$ such that $\sup_{x \in \mathcal{X}} k_i(x, x) \leq \kappa$ and $\text{trace}(\mathbf{L}) \leq \tau$, we have*

$$\hat{R}_l(\mathcal{H}_\lambda^p) \leq \lambda m \sqrt{\frac{\kappa \tau}{l}}.$$

*(B) If $p$ is such that $q \in \mathbb{N}$, where $\frac{1}{p} + \frac{1}{q} = 1$, then*

$$\hat{R}_l(\mathcal{H}_\lambda^p) \leq \frac{\lambda}{l}\sqrt{\eta_0 q}||\mathbf{u}||_q,$$

*where $\eta_0 = \frac{23}{22}$. For separable kernels,*

$$\hat{R}_l(\mathcal{H}_\lambda^p) \leq \lambda m^{1/q}\sqrt{\frac{\eta_0 q \kappa \tau}{l}}.$$

*(C) If $p = 1$, so that $q = \infty$, then*

$$\hat{R}_l(\mathcal{H}_\lambda^1) \leq \frac{\lambda}{l}\sqrt{\eta_0 r}||\mathbf{u}||_r, \forall r \in \mathbb{N}.$$

*For separable kernels, we have*

$$\hat{R}_l(\mathcal{H}_\lambda^1) \leq \begin{cases} \lambda\sqrt{\frac{\eta_0 \kappa \tau}{l}}, & \text{if } m = 1, \\ \lambda\sqrt{\frac{\eta_0 e \lceil 2 \ln m \rceil \kappa \tau}{l}}, & \text{if } m > 1. \end{cases}$$

Due to space limitations, the full proof of this Theorem is provided our Supplementary Material [34].

Using well-known results [7], these bounds on Rademacher complexity can be immediately turned into generalization bounds for our algorithms.

## 4 Empirical Studies

**Statistical Benefits of Joint Input/Output Kernel Learning**: We start with a small dataset of weekly log returns of 9 stocks from 2004, studied in [39, 31] in the context of linear multivariate regression with output covariance estimation techniques. We consider first-order vector autoregressive (VAR) models of the form $\mathbf{x}_t = f(\mathbf{x}_{t-1})$ where $\mathbf{x}_t$ corresponds to the 9-dimensional vector of log-returns for the 9 companies at week $t$ and the function $f$ is estimated by solving Eqn. 6. Our experimental prototcol is exactly the same as [39, 31]: data is split evenly into a training and a test set and the regularizaton parameter $\lambda$ is chosen by 10-fold cross-validation. All other parameters are left at their default values (i.e., $p = 1$). We generated a dictionary of 117 Gaussian kernels defined by univariate Gaussian kernels on each of the 9 dimensions with 13 varying bandwidths. Results are shown in Table 1 where we compare our methods in terms of mean test RMSE against standard linear regression (OLS) and linear Lasso independently applied to each output coordinate, and the sparse multivariate regression with covariance estimation approaches of [31, 39], labeled MRCE and FES respectively. *We see that joint input and output kernel learning (labeled IOKL) yields the best return prediction model reported to date on this dataset.* As expected, it outperforms models obtained by leaving output kernel matrix fixed as the identity and only optimizing scalar kernels (IKL), or only optimizing the output kernel for fixed choices of scalar kernel (OKL). Of the 117 kernels, 13 have 97% of the mass in the learnt scalar kernel combination.

**Table 1:** VAR modeling on financial datasets.

|  | OLS | Lasso | MRCE | FES | IKL | OKL | IOKL |
|---|---|---|---|---|---|---|---|
| WMT | 0.98 | 0.42 | 0.41 | 0.40 | 0.43 | 0.43 | 0.44 |
| XOM | 0.39 | 0.31 | 0.31 | 0.29 | 0.32 | 0.31 | 0.29 |
| GM | 1.68 | 0.71 | 0.71 | 0.62 | 0.62 | 0.59 | **0.47** |
| Ford | 2.15 | 0.77 | 0.77 | 0.69 | 0.56 | 0.48 | **0.36** |
| GE | 0.58 | 0.45 | 0.45 | 0.41 | 0.41 | 0.40 | **0.37** |
| COP | 0.98 | 0.79 | 0.79 | 0.79 | 0.81 | 0.80 | **0.76** |
| Ctgrp | 0.65 | 0.66 | 0.62 | 0.59 | 0.66 | 0.62 | **0.58** |
| IBM | 0.62 | 0.49 | 0.49 | 0.51 | 0.47 | 0.50 | **0.42** |
| AIG | 1.93 | 1.88 | 1.88 | 1.74 | 1.94 | 1.87 | 1.79 |
| Average | 1.11 | 0.72 | 0.71 | 0.67 | 0.69 | 0.67 | **0.61** |

**Scalability and Numerical Behaviour**: Our main interest here is to observe the classic tradeoff in numerical optimization between running few, but very expensive steps versus executing several cheap iterations. We use a 102-class image categorization dataset – Caltech-101 – which has been very well studied in the multiple kernel learning literature [12, 37, 14]. There are 30 training images per category for a total of 3060 training images, and 1355 test images. Targets are 102-dimensional class indicator vectors. We define a dictionary of kernels using 10 scalar-valued kernels precomputed from visual features and made publically available by the authors of [37], for 3 training/test splits. From previous studies, it is well known that all underlying visual features contribute to object discrimination on this dataset and hence non-sparse multiple kernel learning with $l_p, p > 1$ norms are more effective. We therefore set $p = 1.7$ and $\lambda = 0.001$ without any further tuning, since their choice is not central to our main goals in this experiment. We vary the stopping criteria for our CG-based Sylvester solver ($cg_\epsilon$) and the number of iterations ($sdp_{iter}$) allowed in the Sparse SDP solver, for the **C** and **L** subproblems respectively. Note that the closed form $\boldsymbol{\eta}$ updates (Eqn. 10) for $l_p$ norms take negligible time.

We compare our algorithms with an implementation in which each subproblem is solved exactly using an eigendecomposition based Sylvester solver for **C**, and unconstrained updates for **L** developed in [12], respectively. To make comparisons meaningful, we set $\tau$ to a large value so that the optimization over $\mathbf{L} \in \mathcal{S}_+^n(\tau)$ effectively corresponds to unconstrained minimization over the entire psd cone $\mathcal{S}_+^n$. In Figure 1, 2, we report the improvement in objective function and classification accuracy as a function of time (upto 1 hour). We see that insufficient progress is made in both extremes: when either the degree of inexactness is intolerable ($cg_\epsilon = 0.1, sdp_{iter} = 100$) or when subproblems are solved to very high precision ($cg_\epsilon = 10^{-6}, sdp_{iter} = 3000$). *Our solvers are far more efficient than eigendecomposition based implementation that takes an exorbitant amount of time per iteration for exact solutions. Approximate solvers at appropriate precision (e.g., $cg_\epsilon = 0.01, sdp_{iter} = 1000$) make very rapid progress and return high accuracy models in just a few minutes*. In fact, averaged over the three training/test splits, *the classification accuracy obtained is $79.43\% \pm 0.67$ which is highly competitive with state of the art results reported on this dataset*, with the kernels used above. For example, [37] report $78.2\% \pm 0.4$, [14] report $77.7\% \pm 0.3$ and [12] report $75.36\%$.

**Figure 1:** Objective function vs time

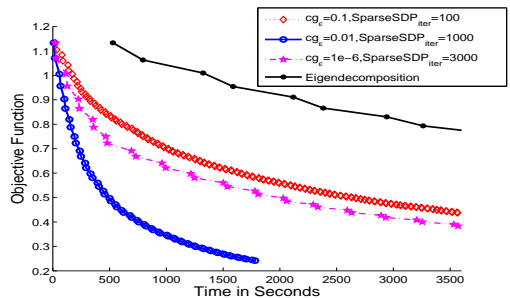

**Figure 2:** Accuracy vs time

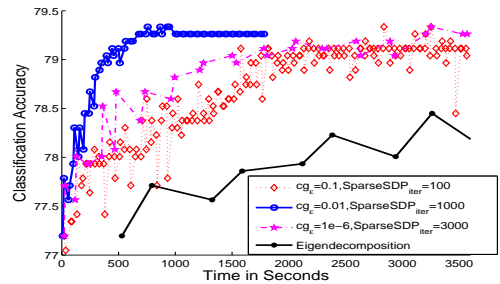

### 4.1 Application: Non-linear Causal Inference

Here, our goal is to show how high-dimensional causal inference tasks can be naturally cast as sparse function estimation problems within our framework, leading to novel nonlinear extensions of Grouped Graphical Granger Causality techniques (see [33, 22] and references therein). In this setting, there is an interconnected system of $N$ distinct sources of high dimensional time series data which we denote as $\mathbf{x}_t^i \in \mathbb{R}^{d_i}, i = 1 \ldots N$. We refer to these sources as "nodes". The system is observed from time $t = 1$ to $t = T$, and the goal is to infer the causal relationships between the nodes. Let $G$ denote the adjacency matrix of the unknown causal interaction graph where $G_{ij} > 0$ implies that node $i$ causally influences node $j$. In 1980, Clive Granger gave an operational definition for Causality:

**Granger Causality** [15]: *A subset of nodes $A_i = \{j : G_{ij} > 0\}$ is said to causally influence node $i$, if the past values of the time series collectively associated with the node subset $A_i$ is predictive of the future evolution of the time series associated with node $i$, with statistical significance, and more so than the past values of $i$ alone.*

A practical appeal of this definition is that it links causal inference to prediction, with the caveat that causality insights are bounded by the quality of the underlying predictive model. Furthermore, the prior knowledge that the underlying causal interactions are highly selective makes sparsity a meaningful prior to use. Prior work on using sparse modeling techniques to uncover causal graphs has focused on linear models [33, 22] while many, if not most, natural systems involve nonlinear interactions for which a functional notion of sparsity is more appropriate.

To apply our framework to such problems, we model the system as the problem of estimating $N$ nonlinear functions: $\mathbf{x}_t^i = f^i\left(\mathbf{x}_t^1, \mathbf{x}_{t-1}^1 \ldots \mathbf{x}_{t-L}^1, \ldots, \mathbf{x}_t^N, \mathbf{x}_{t-1}^N \ldots \mathbf{x}_{t-L}^N\right)$, for $1 \leq i \leq N$, and where $L$ is a lag parameter. The dynamics of each node, $f^i$, can be expressed as the sum of a set of vector-valued functions,

$$f^i = \sum_{j=1,s}^{N} f_{j,s}^i \qquad (14)$$

where the component $f_{j,s}^i$, for all values of the index $s$, *only* depends on the history of node $j$, i.e., the observations $\mathbf{x}_{t-1}^j \ldots \mathbf{x}_{t-L}^j$. Each $f_{j,s}^i$ belongs to vector-valued RKHS whose kernel is $k_{j,s}(\cdot,\cdot)\mathbf{L}^i$. In other words, we set up a dictionary of separable matrix-valued kernels $\mathcal{D}_{\mathbf{L}^i} = \{k_{j,s}\mathbf{L}^i\}_{j,s}$, where scalar kernels $k_{j,s}$ depend only on individual nodes $j$ alone; and the output matrix $\mathbf{L}^i$ is associated with node $i$ currently being modeled. By imposing (functional) sparsity in the sum in Eqn. 14 using our framework, i.e. estimating $f^i$ by solving Eqn. 6, we can identify which subset of nodes are causal drivers (in the Granger sense) of the dynamics observed at node $i$. The sparsity structure of $f^i$ then naturally induces a weighted causal graph $G$:

$$G_{ij} = \sum_{s} \eta_{j,s}^i$$

where $\eta_{j,s}^i$ are the kernel weights estimated by our algorithm. Note that $G_{ij} \neq 0$ only if a component function associated with the history of node $j$ (for some $s$) is non-zero in the sum Eqn. 14). In addition to recovering the temporal causal interactions in this way, the estimated output kernel matrix $\mathbf{L}^i$ associated with each $f^i$ captures within-source temporal dependencies. We now apply these ideas to a problem in computational biology.

**Causal Inference of Gene Networks**: We use time-course gene expression microarray data measured during the full life cycle of Drosophila melanogaster [2]. The expression levels of 4028 genes are simultaneously measured at 66 time points corresponding to various developmental stages. We extracted time series data for 2397 unique genes, and grouped them into 35 functional groups based on their gene ontologies. The goal is to infer causal interactions between functional groups (represented by multiple time series associated with genes in that group), as well obtain insight on within-group relationships between genes. We conducted four sets of experiments: with linear and nonlinear dictionaries (Gaussian kernels with 13 choices of bandwidths per group), and with or without output kernel learning. We use the parameters $\lambda = 0.001$ and time lag of 7 without tuning. Figure 3 shows holdout RMSE from the four experiments, for each of the 35 functional groups. Clearly, nonlinear models with both input and output kernel learning (labeled "nonlinear L" in Figure 3) give the best predictive performance implying greater relability in the implied causal graphs.

**Figure 3:** RMSE in predicing multiple time series

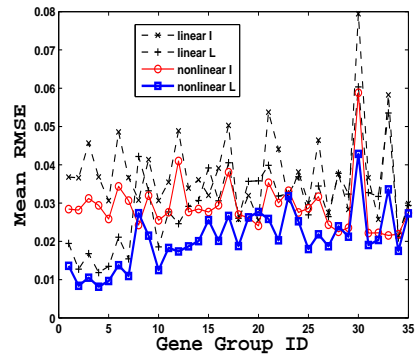

In consultation with a professional biologist, we analyzed the causal graphs uncovered by our approach (Figure 4). In particular the nonlinear causal model uncovered the centrality of a key cellular enzymatic activity, that of helicase, which was not recognized by the linear model. In contrast, the central nodes in the linear model are related to membranes (lipid binding and gtpase activity). Nucleic acid binding transcription factor activity and transcription factor binding are both related to the helicase activity, which is consistent with biological knowledge of them being tightly coupled. This was not captured in the linear model. Molecular chaperone functions, which connect ATPase activity and unfolded protein binding, was successfully identified by our model, while the linear model failed to recognize its relevance. It is less likely that unfolded protein and lipid activity should be linked as suggested by the linear model.

**Figure 4:** Causal Graphs: Linear (left) and Non-linear (right)

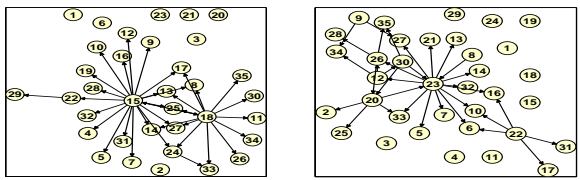

In addition, via output kernel matrix estimation (i.e., $\mathbf{L}^i$), our model also provides insight on the conditional dependencies within genes, shown in Figure 5, for the *unfolded protein binding* group.

**Figure 5:** Interactions in *unfolded protein binding* group

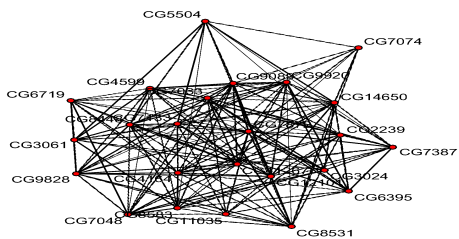

## 5 Related work and Conclusion

Our work is the first to address efficient simultaneous estimation of both the input and output components of separable matrix-valued kernels. Two recent papers are closely related. In [12], the input scalar kernel is predefined and held fixed, while the output matrix is optimized in a block coordinate descent procedure. As discussed in Section 2, this approach involves solving Sylvester equations using eigendecomposition methods which is computationally very costly. In very recent work, concurrent with our work, [18] independently propose a multiple kernel learning framework for operator-valued kernels. Some elements of their work are similar to ours. However, they only optimize the scalar input kernel keeping the output matrix fixed. Their optimization strategy also includes eigendecomposition, and Gauss-Siedel iterations for solving linear systems, while we exploit the quadratic nature of the objective function using CG and a fast sparse SDP solver to demonstrate the scalability benefits of inexact optimization. In addition, we provide generalization analysis in terms of bounds on the Rademacher complexity of our vector-valued hypothesis spaces, complementing analogous results in the scalar multiple kernel learning literature [11, 20]. We also outlined how our framework operationalizes nonlinear Granger Causality in high-dimensional time series modeling problems, which may be of independent interest. Future work includes extending our framework to other classes of vector-valued kernels [1, 9] and to functional data analysis problems [30].

**Acknowledgments:** We thank Haim Avron, Satyen Kale, Rick Lawrence, Bonnie Ray and Tara Sainath for several technically insightful and enthusiastic conversations on this work.